\documentclass[11pt, a4paper]{article}

\usepackage[margin=1in]{geometry}

\usepackage{mathptmx} 
\usepackage[T1]{fontenc}

\usepackage{authblk}

\usepackage{graphicx}
\usepackage{hyperref}
\usepackage{xcolor}
\usepackage{booktabs} 
\usepackage{amsmath}
\usepackage{cleveref}
\usepackage{float}

\hypersetup{
    colorlinks=true,
    linkcolor=blue,
    filecolor=magenta,      
    urlcolor=cyan,
    citecolor=blue,
}

\makeatletter
\renewcommand{\maketitle}{
    \begin{center}
        \vspace*{0.5cm}
        \hrule height 1.5pt \vspace{0.4cm}      
        {\huge \textbf{\@title} \par}           
        \vspace{0.4cm} \hrule height 1.5pt      
        \vspace{0.5cm}
        {\large \textsc{A Preprint} \par}       
        \vspace{0.8cm}
        {\large \@author \par}                  
        \vspace{0.5cm}
        {\large \@date \par}                    
    \end{center}
    \vspace{0.5cm}
}
\makeatother

\title{Replace, Don’t Expand: Mitigating Context Dilution in Multi-Hop RAG via Fixed-Budget Evidence Assembly}

\usepackage{authblk}

\newbox{\orcid}
\sbox{\orcid}{\includegraphics[scale=0.06]{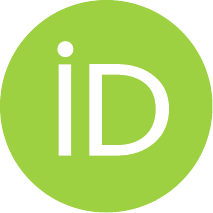}} 

\author[1]{%
    Moshe Lahmy\thanks{\texttt{moshe.lahmi@msmail.ariel.ac.il}}%
}

\author[1,2]{%
    \href{https://orcid.org/0000-0001-5180-4541}{\usebox{\orcid}\hspace{1mm}Roi Yozevitch\thanks{\texttt{roiyo@Ariel.ac.il}}}%
}

\affil[1]{Department of Electrical Engineering, Ariel University, Ariel, Israel}
\affil[2]{Department of Computer and Software Engineering, Ariel University, Ariel, Israel}

\begin{document}

\maketitle

\begin{abstract}
\noindent Retrieval-Augmented Generation (RAG) systems often fail on multi-hop queries when the initial retrieval misses a bridge fact. Prior corrective approaches, such as Self-RAG, CRAG, and Adaptive-$k$, typically address this by \textit{adding} more context or pruning existing lists. However, simply expanding the context window often leads to \textbf{context dilution}, where distractors crowd out relevant information. We propose \textbf{SEAL-RAG}, a training-free controller that adopts a \textbf{``replace, don't expand''} strategy to fight context dilution under a fixed retrieval depth $k$. SEAL executes a (\textbf{S}earch $\rightarrow$ \textbf{E}xtract $\rightarrow$ \textbf{A}ssess $\rightarrow$ \textbf{L}oop) cycle: it performs on-the-fly, entity-anchored extraction to build a live \textit{gap specification} (missing entities/relations), triggers targeted micro-queries, and uses \textit{entity-first ranking} to actively swap out distractors for gap-closing evidence. We evaluate SEAL-RAG against faithful re-implementations of Basic RAG, CRAG, Self-RAG, and Adaptive-$k$ in a shared environment on \textbf{HotpotQA} and \textbf{2WikiMultiHopQA}. On HotpotQA ($k=3$), SEAL improves answer correctness by \textbf{+3 to +13 pp} and evidence precision by \textbf{+12 to +18 pp} over Self-RAG. On 2WikiMultiHopQA ($k=5$), it outperforms Adaptive-$k$ by \textbf{+8.0 pp} in accuracy and maintains \textbf{96\%} evidence \textbf{precision}, compared to 22\% for CRAG. These gains are statistically significant ($p<0.001$). By enforcing fixed-$k$ replacement, SEAL yields a predictable cost profile while ensuring the top-$k$ slots are optimized for precision rather than mere breadth. We release our code and data at \url{https://github.com/mosherino/SEAL-RAG}.

\vspace{1em} 
\noindent\textbf{Keywords:} Retrieval-Augmented Generation; multi-hop reasoning; context dilution; evidence assembly; SEAL-RAG; HotpotQA; 2WikiMultiHopQA
\end{abstract}

\section{Introduction}

Large Language Models (LLMs) augmented with retrieval (RAG) are now the standard for knowledge-intensive tasks\cite{lewis2020rag}. However, standard ``Retrieve-then-Read'' pipelines are brittle in multi-hop scenarios: if the initial top-$k$ retrieval misses a crucial ``bridge'' entity or relation, the generator hallucinates or fails \cite{karpukhin2020dpr, trivedi2020-multihop-condition}.

To mitigate this, recent research has focused on \textbf{iterative and corrective RAG}. Systems like \textbf{Self-Reflective RAG (Self-RAG)} \cite{asai2024selfrag} and \textbf{Corrective RAG (CRAG)} \cite{yan2024crag} introduce feedback loops: they critique the retrieved evidence and trigger additional retrieval steps if gaps are detected. While effective at improving recall, these methods typically operate via \textbf{Breadth-First Addition}: they append new passages to the existing context. This approach assumes that ``more context is better,'' but in production environments with strict latency and token budgets, this assumption fails. Expanding the context window introduces \textbf{distractors} (irrelevant passages that confuse the model) and \textbf{lateral redundancy} (duplicate information), often degrading the model's ability to reason over the specific bridge facts required—a phenomenon known as \textit{context dilution} \cite{liu2023lostinthemiddle}.

We argue that under a fixed inference budget, the goal of a RAG controller should not be to \textit{accumulate} evidence, but to \textit{optimize the composition} of the fixed-$k$ set. We propose a fundamental shift from \textit{retrieval expansion} to \textbf{Fixed-Budget Evidence Assembly}.

\subsection{The SEAL-RAG Approach}

\begin{figure}[ht]
  \centering
  \includegraphics[width=\textwidth]{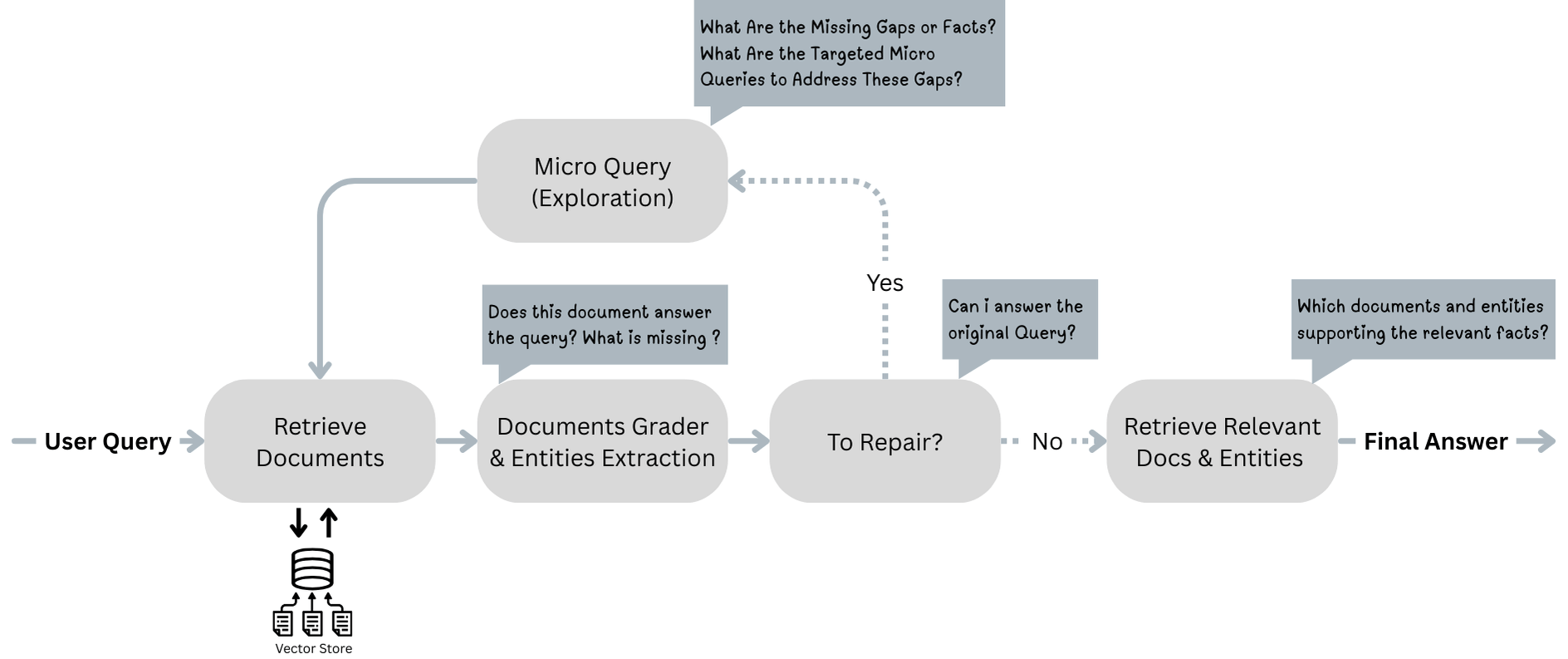}
  \caption{\textbf{SEAL-RAG pipeline (Search $\rightarrow$ Extract $\rightarrow$ Assess $\rightarrow$ Loop).} From a user query, initial retrieval (fixed top-$k$) pulls candidates. Each loop: \textbf{Extract} performs entity-first extraction to form a \textit{gap specification}; \textbf{Assess} applies scope-aware sufficiency to decide \textit{stop} vs.\ \textit{repair}. On repair, the \textit{Micro-Query} policy explores targeted queries. New evidence is integrated via \textit{entity-first ranking} to replace distractors; once sufficient, the system emits the answer.}
  \label{fig:seal_pipeline}
\end{figure}

We introduce \textbf{SEAL-RAG}, a training-free inference-time controller designed for \textbf{Fixed-$k$ Gap Repair}. Unlike prior methods that treat the context window as an append-only log, SEAL treats the top-$k$ slots as a scarce resource. The core mechanism is a (\textbf{S}earch $\rightarrow$ \textbf{E}xtract $\rightarrow$ \textbf{A}ssess $\rightarrow$ \textbf{L}oop) cycle (\Cref{fig:seal_pipeline}). Instead of relying on implicit scalar confidence scores, SEAL performs on-the-fly entity extraction to build an \textbf{Explicit Gap Specification} (e.g., ``Missing the \textit{founding date} of \textit{Organization X}''). It translates these gaps into targeted micro-queries and employs an \textbf{Entity-First Replacement} policy: new candidates are scored based on their ability to close specific gaps and are used to \textit{evict} the lowest-utility passages (distractors) from the current set. This maintains a constant context size ($k$) while strictly increasing information density.

\subsection{Contributions}

This work makes the following contributions:

\begin{itemize}
    \item \textbf{Controller-Level Framing: Fixed-Budget, Gap-Aware Evidence Repair.}
    We recast multi-hop RAG as a constrained \emph{evidence set optimization} problem:
    the system must maintain a small evidence set $E$ of fixed size $k$ that is sufficient
    to answer the query. While primitives such as entity extraction, relation extraction,
    and micro-queries are well-established, prior work typically uses them to
    \emph{expand} a candidate pool and then select from an ever-growing or static context.
    In contrast, SEAL-RAG introduces a controller that (i) maintains an explicit
    \emph{gap specification} over an entity ledger (tracking which entities/relations are
    supported or missing), and (ii) uses this specification to drive a
    \emph{replacement-based repair policy} under a fixed-$k$ budget. This
    fixed-budget, gap-aware repair view directly targets the ``context dilution'' failure
    mode of add-only pipelines. All underlying primitives are standard; the contribution lies in the controller's design and in how it combines these primitives.

    \item \textbf{Replace-Not-Expand Mechanics via Entity-Centric Utility.}
    SEAL-RAG treats the $k$ context slots as a scarce resource rather than a buffer to be
    filled. At each loop, candidate passages are scored using an entity-centric utility
    function that balances gap coverage, corroboration, novelty, and redundancy.
    Low-utility distractors are actively \emph{evicted} and replaced by candidates that
    better resolve identified gaps, while the ledger and gap specification are updated.
    This combination of (a) targeted micro-queries sourced from explicit gaps and
    (b) active replacement under a strict size constraint distinguishes SEAL-RAG from
    both add-only controllers (e.g., CRAG, Self-RAG) and prune-from-a-static-pool
    selectors (e.g., Adaptive-$k$).
    
    \item \textbf{Unified, Controller-Focused Evaluation.}
    To isolate the effect of controller logic from model training or architectural
    differences, we re-implement the control policies of \textbf{Self-RAG},
    \textbf{CRAG}, and \textbf{Adaptive-$k$} in a shared, training-free environment
    with the \emph{same} retriever, index, and generator. This experimental design
    enables a fair comparison of add-only, prune-only, and repair-based controllers
    under identical retrieval and generation conditions.

    \item \textbf{Empirical Gains on Multi-Hop Benchmarks.}
    We evaluate SEAL-RAG on \textbf{HotpotQA} and \textbf{2WikiMultiHopQA} across
    retrieval depths $k \in \{1,3,5\}$. SEAL-RAG consistently maintains higher
    evidence precision and improves answer accuracy over baselines. For example,
    on 2WikiMultiHopQA at $k=5$, SEAL-RAG attains \textbf{96\%} evidence precision
    compared to \textbf{22\%} for CRAG, and yields an answer accuracy improvement of
    +8.0 percentage points over Adaptive-$k$.
\end{itemize}

\subsection{Paper Organization}
\Cref{sec:related} reviews related work. \Cref{sec:method} details SEAL-RAG (loop controller, scope-aware sufficiency, loop-adaptive extraction, entity-first ranking, and the micro-query policy). \Cref{sec:experiments} specifies datasets, models, retrieval/indexing, baseline, metrics/judging, and protocol. \Cref{sec:results} presents main results at $k=1$, $k=3$, and $k=5$ with per-backbone tables and discussion. \Cref{sec:ablations} reports loop-budget ablations and analysis. \Cref{sec:limitations} states limitations and threats to validity. \Cref{sec:conclusion} concludes the paper. Detailed prompts and statistical tables appear in the Appendix. All code and datasets are available in the GitHub repository at \url{https://github.com/mosherino/SEAL-RAG}.

\section{Related Work}\label{sec:related}

\subsection{Standard RAG and Multi-Hop Challenges}
RAG has evolved from early sparse retrieval pipelines to sophisticated dense and hybrid systems \cite{lewis2020rag, gao2023survey}. While dense retrievers like \textbf{Dense Passage Retrieval (DPR)} \cite{karpukhin2020dpr} and late-interaction models like \textbf{ColBERT} \cite{khattab2020colbert} improve recall on single-hop queries, they often struggle with multi-hop reasoning, where the answer depends on composing information from multiple disjoint documents \cite{min2019compositional, chen2019understanding}. In these scenarios, the standard \textit{retrieve-then-generate} pattern faces a critical bottleneck: if the initial top-$k$ set misses a ``bridge'' fact, the generator cannot recover. A common workaround is to blindly increase $k$ or accumulate more context, but this introduces noise. Recent analysis confirms that irrelevant context can significantly degrade model performance—a phenomenon known as ``lost in the middle'' or \textbf{context dilution} \cite{liu2023lostinthemiddle}. SEAL-RAG targets this specific failure mode by holding $k$ fixed and iteratively \textit{repairing} the evidence set rather than expanding it.

\subsection{Corrective and Reflective RAG}
To mitigate retrieval failures, recent research has shifted towards \textbf{Active Retrieval} \cite{jiang2023active}, where the model actively interacts with the search engine during inference. \textbf{Self-RAG} \cite{asai2024selfrag} integrates retrieval and critique via special reflection tokens, allowing the model to self-assess generation quality and trigger additional retrieval steps when necessary. Similarly, \textbf{CRAG} \cite{yan2024crag} employs a lightweight evaluator to detect low-quality retrieval and trigger corrective actions, such as web searches. While these methods improve robustness against irrelevant context \cite{yoran2023making}, they typically operate via \textbf{Breadth-First Addition}: they append new passages to the existing context window. This approach assumes that ``more context is better,'' but in production environments with strict latency and token budgets, it leads to unbounded context growth and variable inference costs. SEAL-RAG adopts the active spirit of these methods but enforces a \textit{replacement} policy to maintain a predictable budget.

\subsection{Adaptive Retrieval and Pruning}
A parallel line of work focuses on dynamic resource allocation to improve efficiency. \textbf{Adaptive Retrieval-Augmented Generation (Adaptive-RAG)} \cite{jeong2024adaptive} functions as a router, classifying query complexity to dynamically select between retrieval-free and retrieval-augmented paths. \textbf{Adaptive-$k$} \cite{taguchi2025adaptivek} and \textbf{Long-Context Bootstrapper (LC-Boost)} \cite{wu2024lcboost} aim to optimize the context window by pruning irrelevant documents from a larger retrieved list or selecting a minimal sufficient subset. While these methods address the efficiency drawback of standard RAG, they are primarily \textit{selectors} or \textit{routers}, not \textit{repairers}. If the initial retrieval pool misses a bridge fact entirely, pruning cannot recover it. In contrast, SEAL-RAG performs \textbf{Active Repair}: it diagnoses specific missing entities (e.g., via on-the-fly extraction) and issues targeted micro-queries to fetch new evidence that was never in the initial pool, replacing low-utility items to improve the set's composition.

\subsection{Contrast with SEAL-RAG}
SEAL-RAG occupies a distinct position in the design space. Unlike Corrective/Reflective methods (Self-RAG, CRAG), it enforces a \textbf{Fixed Capacity} to prevent context dilution. Unlike Adaptive methods (Adaptive-$k$), it performs \textbf{Active Repair} via targeted micro-queries rather than passive pruning. By combining explicit gap modeling with a replacement policy, SEAL optimizes the \textit{composition} of the top-$k$ slots, ensuring high precision under strict budget constraints.

\section{SEAL-RAG (Method)}\label{sec:method}

\subsection{Problem Formulation \& Architecture}
We formalize retrieval-augmented generation under strict budgets as a constrained set-optimization problem. Given a query $q$ and a corpus $\mathcal{C}$, our goal is to identify an optimal evidence set $E^* \subset \mathcal{C}$ that maximizes the probability of generating a correct answer $a$, subject to a cardinality constraint $|E| = k$. In this framework, we define ``budget'' strictly as the finite context capacity ($k$) available to the generator, treating the evidence window as a scarce cognitive resource to be optimized rather than merely a computational cost to be minimized.

Unlike standard RAG, which approximates $E^*$ via a single retrieval pass, or corrective methods that relax the constraint (allowing $|E| > k$), SEAL-RAG iteratively refines $E$ while strictly enforcing $|E_t| = k$ at every step $t$.

\begin{figure}[ht!]
  \centering
  \includegraphics[width=0.5\linewidth]{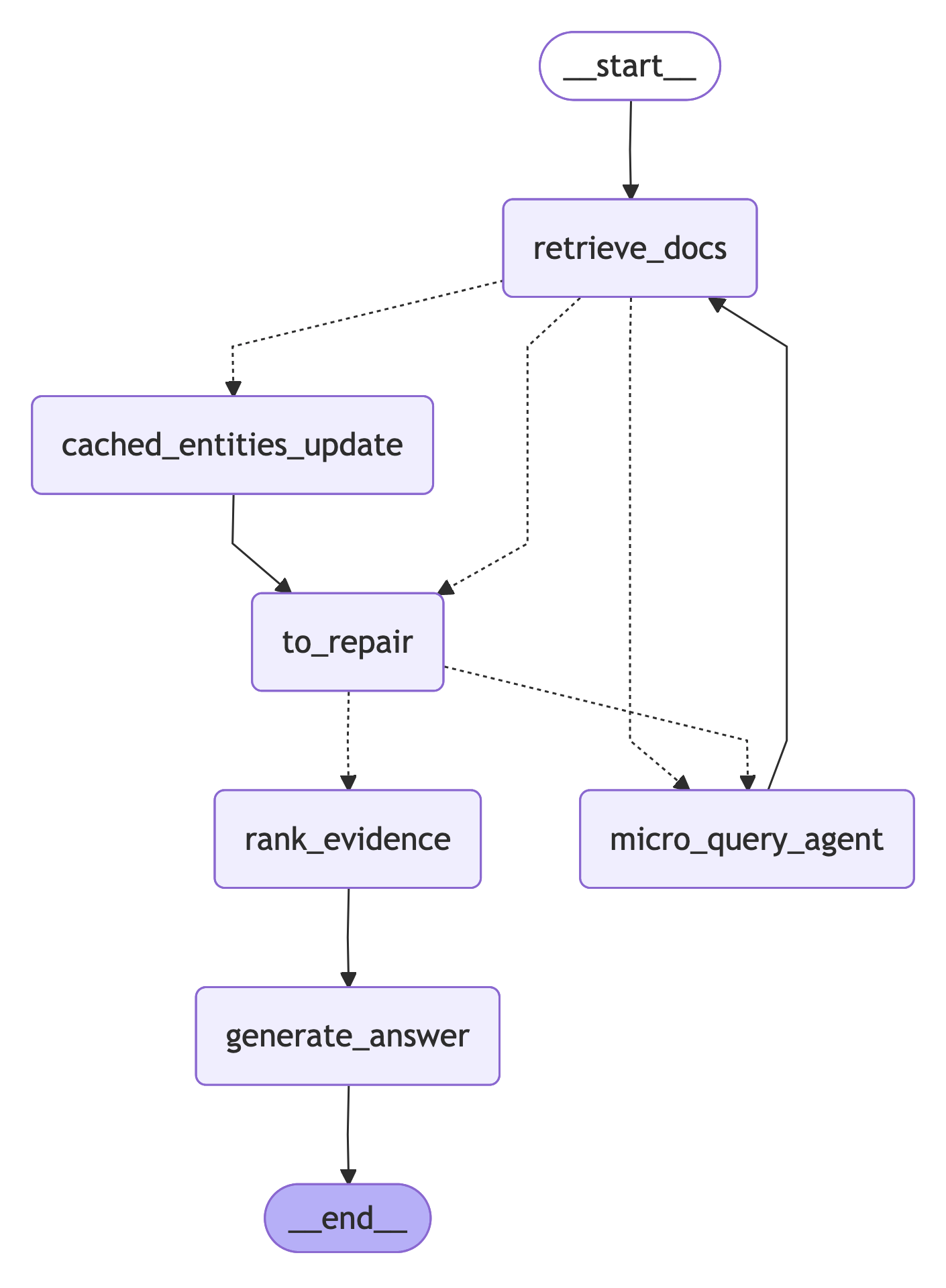}
  \caption{\textbf{Execution graph for SEAL-RAG.} Nodes represent logical stages: \texttt{retrieve\_docs} initializes the fixed-$k$ set; \texttt{cached\_entities\_update} builds the ledger $U_t$; \texttt{to\_repair} acts as the sufficiency gate. If repair is needed, \texttt{micro\_query\_agent} fetches candidates and \texttt{rank\_evidence} performs replacement. Solid arrows denote the primary path; dashed arrows indicate loopbacks.}
  \label{fig:execution_graph}
\end{figure}

The controller maintains a state tuple $S_t = (E_t, U_t, B_t)$, where:
\begin{itemize}
    \item $E_t$: The current evidence buffer of fixed size $k$.
    \item $U_t$: A structured \textit{Entity Ledger} derived from $E_t$ (containing entities, relations, and provenance).
    \item $B_t$: A \textit{Blocklist} of unproductive query patterns or sources to prevent cycles.
\end{itemize}

The inference process follows a (\textbf{S}earch $\rightarrow$ \textbf{E}xtract $\rightarrow$ \textbf{A}ssess $\rightarrow$ \textbf{L}oop) cycle (\Cref{fig:execution_graph}). \textbf{Initialization:} At $t=0$, $E_0$ is populated via a standard dense/hybrid retrieval pass. At each subsequent step, the controller assesses sufficiency; if insufficient, it executes a repair policy $\pi(S_t)$ to replace low-utility items in $E_t$, halting when sufficiency is met or a loop budget $L$ is exhausted.

\subsection{State Representation: The Entity Ledger}
To make the ``stop vs.\ repair'' decision computable, SEAL-RAG projects the unstructured evidence $E_t$ into a structured \textit{Entity Ledger} $U_t$. We employ a lightweight, on-the-fly extraction module grounded in Open Information Extraction principles \cite{angeli-etal-2015-leveraging, bhardwaj-etal-2019-carb}.

The extraction process enforces a \textbf{Verbatim Constraint}: extracted facts must be explicitly supported by text spans in $E_t$ to prevent hallucination. The ledger $U_t$ tracks:
\begin{itemize}
    \item \textbf{Entities \& Aliases:} Canonical entities (e.g., ``Theresa May'') mapped to surface forms (``PM May'', ``She'').
    \item \textbf{Typed Relations:} Triplets $(h, r, t)$ linking entities (e.g., \texttt{(Theresa May, authored, Article 50 letter)}).
    \item \textbf{Qualifiers:} Critical metadata such as dates, locations, or roles attached to relations (e.g., \texttt{date=2017}).
\end{itemize}
This structured view allows the controller to detect partial coverage (e.g., the relation exists, but the date qualifier is missing).

\subsection{Sufficiency Assessment}
The \textbf{Sufficiency Gate} evaluates a predicate $\text{Suff}(q, U_t)$ based on four aggregated signals. We employ the LLM as a zero-shot estimator to score these components:
\begin{itemize}
    \item \textbf{Coverage:} The fraction of required question attributes (derived from the query schema) currently present in $U_t$.
    \item \textbf{Corroboration:} The degree of multi-source agreement for critical facts.
    \item \textbf{Contradiction:} Detection of conflicting attribute values across passages.
    \item \textbf{Answerability:} A calibrated confidence score estimating if the question is answerable given $U_t$ \cite{rajpurkar2018squadv2}.
\end{itemize}
If $\text{Suff}(q, U_t)$ is true, the loop terminates. If false, the controller proceeds to gap diagnosis.

\subsection{Gap-Driven Retrieval Policy}
When sufficiency fails, standard corrective methods often rely on generic query rewriting. In contrast, SEAL-RAG computes an \textbf{Explicit Gap Specification} $G_t = \mathcal{N}(q) \setminus U_t$. The system parses the question $q$ to identify necessary information needs $\mathcal{N}(q)$ (e.g., ``Need: Birthplace of Person X'') and subtracts the facts already present in $U_t$.

We categorize gaps into three types:
\begin{enumerate}
    \item \textbf{Missing Entity:} A bridge entity referenced by a relation is absent (e.g., ``The band that released \textit{Parklife}'').
    \item \textbf{Missing Relation:} Two entities are known, but the link between them is unproven.
    \item \textbf{Missing Qualifier:} A relation is known, but a required date or location is missing.
\end{enumerate}

The controller translates $G_t$ into \textbf{atomic micro-queries} (e.g., ``Blur band Parklife release year''). This is significantly more precise than a broad rewrite (e.g., ``Tell me about Blur and Parklife''), which often retrieves general biography pages rather than the specific missing attribute. This policy minimizes query drift and ensures that retrieved candidates are semantically aligned with the specific missing link \cite{carpineto2012survey}. To prevent cycles, the policy updates the blocklist $B_t$ with query terms that failed to yield novel information.

\subsection{Fixed-Capacity Replacement}
The core innovation of SEAL-RAG is its refusal to expand the context window. We treat evidence assembly as a \textbf{Budgeted Maximization} problem: given a candidate pool $C_t$ retrieved via micro-queries, we must select a subset to replace low-utility items in $E_t$ such that the total size remains $k$.

We define an \textbf{Entity-First Utility} function $S(c | U_t)$ to score each candidate $c \in C_t$. Inspired by Maximal Marginal Relevance (MMR) \cite{carbonell1998mmr}, this score balances relevance against redundancy:
\begin{equation}
    S(c) = \lambda_1 \cdot \text{GapCov}(c, G_t) + \lambda_2 \cdot \text{Corr}(c, U_t) + \lambda_3 \cdot \text{Nov}(c, U_t) - \lambda_4 \cdot \text{Red}(c, E_t)
\end{equation}
where $\lambda_i$ are hyperparameters weighting the components:
\begin{itemize}
    \item \textbf{GapCov}: Measures if candidate $c$ contains the specific missing entity or relation defined in the gap set $G_t$.
    \item \textbf{Corr}: Rewards candidates that corroborate existing uncertain facts in the ledger $U_t$ (increasing confidence).
    \item \textbf{Nov}: Rewards non-lateral novelty (introducing new entities or relations not yet in $U_t$).
    \item \textbf{Red}: Penalizes lexical overlap with existing passages in $E_t$ to prevent lateral redundancy.
\end{itemize}

To update the set, the controller identifies the lowest-scoring victim $v \in E_t$ and the highest-scoring candidate $c^* \in C_t$. If $S(c^*) > S(v) + \epsilon$, a swap occurs: $E_{t+1} \leftarrow (E_t \setminus \{v\}) \cup \{c^*\}$. The term $\epsilon$ is a small hysteresis threshold to prevent thrashing (replacing an item with a marginally better one, wasting a loop step). Additionally, we enforce a \textbf{Dwell-Time Guard}: newly inserted items are protected from eviction for one iteration to ensure they are processed by the sufficiency gate before being discarded. This ensures that the information density of the top-$k$ slots strictly increases, actively fighting context dilution.

\subsection{Complexity \& Budget}
A critical advantage of SEAL-RAG is its predictable cost profile. Let $L$ be the maximum loop budget and $k$ the fixed retrieval depth. The total inference cost is bounded by:
\begin{equation}
    \text{Cost}_{\text{SEAL}} = O(L \cdot \text{Retriever}) + O(L \cdot \text{Extractor}) + O(1 \cdot \text{Generator}_k)
\end{equation}
Since the generator is invoked only once on a fixed context of size $k$, the expensive decoding step remains constant regardless of the number of repair loops. In contrast, addition-based methods increase the context size at every step, causing the generator cost to grow super-linearly with $L$. SEAL-RAG guarantees that latency and token usage remain within a tight, pre-calculated envelope $O(k \cdot L)$.

\section{Experimental Setup}\label{sec:experiments}

\subsection{Datasets}
We evaluate on two multi-hop QA benchmarks to assess performance and generalization across different reasoning types.
\begin{itemize}
    \item \textbf{HotpotQA (Distractor Setting):} We use a seeded validation slice of $N=1{,}000$ questions \cite{yang2018hotpotqa}. This dataset primarily tests \textit{bridge} reasoning (connecting entity A to entity B) and \textit{comparison} reasoning (e.g., ``Who is older, X or Y?'').
    \item \textbf{2WikiMultiHopQA:} To assess robustness beyond HotpotQA, we evaluate on a seeded slice of $N=200$ examples from the 2WikiMultiHopQA validation set \cite{ho2020constructing}. This dataset involves complex \textit{compositional} reasoning and inference rules over Wikipedia entities.
\end{itemize}
For both datasets, we use a fixed random seed to ensure deterministic sampling. Representative examples of these reasoning types and how SEAL-RAG handles them are provided in \cref{sec:appendix_examples}.

\subsection{Shared Environment}
To isolate the contribution of the controller logic, we enforce a strict \textbf{Shared Environment}. All methods (SEAL-RAG and baselines) are implemented as workflows using \textbf{LangGraph} \cite{langchain2024langgraph} to ensure consistent state management. They share the exact same underlying components:
\begin{itemize}
    \item \textbf{Indexing Pipeline:} We employ \textbf{Natural Document Segmentation}. Instead of arbitrary sliding windows, we concatenate the page title and all associated sentences provided by the benchmark into a single retrieval unit. This preserves the semantic integrity of documents. Full indexing details are provided in \cref{sec:appendix_implementation}.
    \item \textbf{Retriever:} Dense retrieval using OpenAI embeddings (\texttt{text-embedding-3-small}) and a Pinecone vector store.
    \item \textbf{Unified Backbone Architecture:} To strictly isolate the algorithmic contribution of the controller logic from latent model capabilities, we employ a unified backbone strategy. Within each experimental configuration, the \textit{same} underlying LLM instance powers both the internal Controller (handling entity extraction, sufficiency estimation, and ranking) and the final Generator. We evaluate across the GPT-4 family to ensure robustness: \texttt{gpt-4o} and \texttt{gpt-4o-mini} on 2WikiMultiHopQA, and \texttt{gpt-4.1} and \texttt{gpt-4.1-mini} on HotpotQA. All model calls utilize temperature 0 to ensure deterministic reproducibility.
\end{itemize}
This setup ensures that any performance difference is attributable solely to the retrieval policy (e.g., replacement vs.\ addition), not to differences in the underlying model, index, or prompt engineering.

\subsection{Baselines}
We compare SEAL-RAG against four baselines, re-implemented in our shared environment to match the specific logic of their original proposals. Detailed graph topologies and system prompts for these re-implementations are provided in \cref{sec:appendix_baselines}.
\begin{itemize}
    \item \textbf{Basic RAG:} A linear \texttt{Retrieve} $\rightarrow$ \texttt{Generate} graph. It retrieves $k$ passages once and generates an answer.
    \item \textbf{Self-RAG:} A reflective graph that grades documents for relevance and generations for hallucinations \cite{asai2024selfrag}. If the generation is unsupported, the system loops back to transform the query (capped at 3 attempts).
    \item \textbf{CRAG:} A corrective graph \cite{yan2024crag}. If retrieved documents are graded as ``irrelevant,'' the system triggers an external web search (via Tavily) to augment the context before generation.
    \item \textbf{Adaptive-$k$:} A dynamic pruning method \cite{taguchi2025adaptivek}. It retrieves a large candidate pool ($k=50$) and selects the optimal cut-off point using the ``Largest Gap'' strategy on similarity scores. We evaluate both \textit{Buffer} and \textit{No-Buffer} variants.
\end{itemize}

\subsection{Metrics and Judging}
We report two primary metrics:
\begin{itemize}
    \item \textbf{Judge-EM (Correctness):} We use \textbf{GPT-4o} as an external judge. The judge evaluates \textit{Factual Consistency} against the ground truth, penalizing contradictions or ``I don't know'' responses if the answer exists \cite{zheng2023judging}. The judge sees only the retrieved passages to prevent parametric leakage.
    \item \textbf{Evidence Quality:} We compute \textbf{Gold-title Precision@$k$} and \textbf{Recall@$k$}. To ensure rigorous evaluation, we apply \textbf{Alias Normalization}: retrieved titles are matched against gold titles using a redirect map (e.g., mapping ``JFK'' to ``John F. Kennedy'') to prevent false negatives.
\end{itemize}
We report statistical significance using McNemar's test for correctness and paired $t$-tests for retrieval metrics ($p < 0.05$).

\section{Results}\label{sec:results}

\subsection{Main Results on HotpotQA (\texorpdfstring{$k=1$}{k=1})}
We first evaluate performance under the strictest constraint: a single retrieval slot ($k=1$). In this regime, the system must identify and retain the single most critical passage (often a bridge entity) to answer correctly. Any distractor in this slot results in immediate failure. \Cref{tab:hotpot_k1} presents the results on the seeded $N=1{,}000$ validation slice.

\begin{table}[h]
\centering
\small
\setlength{\tabcolsep}{5pt}
\caption{\textbf{Main results at fixed $k=1$ on HotpotQA ($N=1{,}000$).} Metrics are percentages. All methods share the same environment (models, vector store, judge, metrics); only control logic differs. $\Delta$ is the Judge-EM gain of SEAL-RAG over the best baseline for the \textit{same model}.}
\label{tab:hotpot_k1}
\begin{tabular}{ll cccc c}
\toprule
& & \multicolumn{4}{c}{\textbf{Metrics (\%)}} & \textbf{$\Delta$ (pp)} \\
\cmidrule(lr){3-6} \cmidrule(lr){7-7}
\textbf{Model} & \textbf{Method} & \textbf{Judge-EM} & \textbf{Prec} & \textbf{Rec} & \textbf{F1} & \textbf{$\Delta$ EM} \\
\midrule
gpt-4o-mini & Basic RAG & 41 & 85 & 42 & 57 & \\
& Self-RAG & 48 & 61 & 31 & 41 & \\
& CRAG & 55 & 42 & 21 & 28 & \\
& \textbf{SEAL-RAG} & \textbf{62} & \textbf{86} & \textbf{44} & \textbf{58} & \textbf{+7} \\
\midrule
gpt-4o & Basic RAG & 41 & 85 & 42 & 57 & \\
& Self-RAG & 59 & 75 & 37 & 50 & \\
& CRAG & 58 & 54 & 27 & 36 & \\
& \textbf{SEAL-RAG} & \textbf{73} & \textbf{91} & \textbf{62} & \textbf{72} & \textbf{+14} \\
\midrule
gpt-4.1-mini & Basic RAG & 39 & 85 & 42 & 57 & \\
& Self-RAG & 49 & 72 & 36 & 48 & \\
& CRAG & 52 & 57 & 29 & 38 & \\
& \textbf{SEAL-RAG} & \textbf{71} & \textbf{87} & \textbf{48} & \textbf{61} & \textbf{+19} \\
\midrule
gpt-4.1 & Basic RAG & 40 & 85 & 42 & 57 & \\
& Self-RAG & 63 & 79 & 40 & 53 & \\
& CRAG & 58 & 57 & 28 & 38 & \\
& \textbf{SEAL-RAG} & \textbf{73} & \textbf{90} & \textbf{66} & \textbf{74} & \textbf{+10} \\
\bottomrule
\end{tabular}
\end{table}

\paragraph{Key Observations.}
\begin{itemize}
    \item \textbf{Replacement is decisive at a single slot.} With only one evidence slot, the ability to \textit{displace} a low-yield passage is pivotal. Across all backbones, SEAL-RAG improves Judge-EM by \textbf{+7 to +19 pp} over the best baseline. This confirms that gap-aware micro-queries reliably surface the one page that actually closes the bridge.
    
    \item \textbf{Precision lift without harming recall.} Gold-title Precision rises for SEAL-RAG versus the strongest baseline in each backbone (e.g., 91 vs.\ 75 for gpt-4o). Crucially, Recall stays comparable or higher (e.g., 66 vs.\ 40 for gpt-4.1), yielding a consistent F1 advantage. This refutes the notion that replacement inherently sacrifices coverage.
    
    \item \textbf{Addition-first underperforms at small $k$.} CRAG and Self-RAG broaden context during the loop, but when the reader is constrained to $k=1$, breadth does not help unless it reorders the final top-1. This is visible where CRAG's recall drops (e.g., 21--29\%), as it may append relevant docs to positions $k>1$ which are then truncated. SEAL's replacement policy ensures the best document lands in the \textit{visible} slot.

    \item \textbf{Shift from Read-Time to Retrieval-Time Reasoning.} The performance gap highlights a structural distinction. Standard RAG relies on \textit{Read-Time Reasoning}, requiring simultaneous access to disjoint evidence (Hop 1 and Hop 2), which is physically impossible at $k=1$. SEAL-RAG shifts this to \textit{Retrieval-Time Reasoning}: the controller resolves the bridge entity into the ledger, effectively ``consuming'' the first hop. This allows the single context slot to be dedicated entirely to the final answer-bearing document, rendering the task solvable.
    
\end{itemize}

\subsection{Main Results on HotpotQA (\texorpdfstring{$k=3$}{k=3})}

We next evaluate performance at $k=3$, a standard setting for production RAG systems. With three slots, the challenge shifts from finding a single needle to assembling a coherent set that covers multiple hops without admitting distractors. \Cref{tab:hotpot_k3} details the results.

\begin{table}[h]
\centering
\small
\setlength{\tabcolsep}{5pt}
\caption{\textbf{Main results at fixed $k=3$ on HotpotQA ($N=1{,}000$).} Even with larger capacity, SEAL-RAG maintains a significant lead in Precision and Correctness. $\Delta$ shows the gain over the best baseline ($p < 0.001$).}
\label{tab:hotpot_k3}
\begin{tabular}{ll cccc c}
\toprule
& & \multicolumn{4}{c}{\textbf{Metrics (\%)}} & \textbf{$\Delta$ (pp)} \\
\cmidrule(lr){3-6} \cmidrule(lr){7-7}
\textbf{Model} & \textbf{Method} & \textbf{Judge-EM} & \textbf{Prec} & \textbf{Rec} & \textbf{F1} & \textbf{$\Delta$ EM} \\
\midrule
gpt-4o-mini & Basic RAG & 63 & 49 & 72 & 59 & \\
& Self-RAG & 60 & 66 & 47 & 53 & \\
& CRAG & 62 & 30 & 36 & 33 & \\
& \textbf{SEAL-RAG} & \textbf{69} & \textbf{84} & \textbf{44} & \textbf{57} & \textbf{+6} \\
\midrule
gpt-4o & Basic RAG & 68 & 49 & 72 & 59 & \\
& Self-RAG & 71 & 76 & 55 & 61 & \\
& CRAG & 69 & 37 & 44 & 40 & \\
& \textbf{SEAL-RAG} & \textbf{77} & \textbf{89} & \textbf{68} & \textbf{75} & \textbf{+6} \\
\midrule
gpt-4.1-mini & Basic RAG & 64 & 49 & 72 & 59 & \\
& Self-RAG & 64 & 73 & 56 & 61 & \\
& CRAG & 67 & 40 & 49 & 43 & \\
& \textbf{SEAL-RAG} & \textbf{77} & \textbf{86} & \textbf{49} & \textbf{61} & \textbf{+10} \\
\midrule
gpt-4.1 & Basic RAG & 68 & 49 & 72 & 59 & \\
& Self-RAG & 73 & 79 & 61 & 66 & \\
& CRAG & 72 & 41 & 50 & 44 & \\
& \textbf{SEAL-RAG} & \textbf{76} & \textbf{91} & \textbf{73} & \textbf{79} & \textbf{+3} \\
\bottomrule
\end{tabular}
\end{table}

\paragraph{Key Observations.}
\begin{itemize}
    \item \textbf{Precision lead persists under larger capacity.} With three slots, recall naturally rises for all methods. However, SEAL-RAG maintains a massive Precision advantage (e.g., \textbf{89\%} vs.\ 37--76\% for gpt-4o). This indicates that while baselines use the extra slots to accumulate near-duplicates or topical distractors, SEAL uses them to store complementary bridge facts.
    
    \item \textbf{Replacement reduces lateral redundancy.} By treating the evidence set as a fixed-capacity buffer, SEAL-RAG actively evicts redundant passages (e.g., two biographies of the same person) to make room for the second hop (e.g., the organization page). This raises Precision without sacrificing the Recall that naturally comes with $k=3$, translating into the highest Judge-EM across all backbones.
    
    \item \textbf{Addition-first recall is offset by distractors.} While CRAG and Basic RAG often achieve high recall (e.g., 72\%), their low precision (30--49\%) drags down answer correctness. This confirms that simply having the answer in the context is insufficient if it is buried in noise; the model requires a \textit{curated} context to reason reliably.
\end{itemize}

\subsection{Generalization to 2WikiMultiHopQA (\texorpdfstring{$k=1, 3, 5$}{k=1, 3, 5})}

To address concerns regarding generalization, we evaluate on \textbf{2WikiMultiHopQA} ($N=200$), which requires complex compositional reasoning. We extend the evaluation to $k=5$ to explicitly test the ``more context is better'' assumption. \Cref{tab:2wiki_stacked} presents the results stacked by retrieval depth.

\begin{table}[h]
\centering
\small
\setlength{\tabcolsep}{4.5pt}
\caption{\textbf{2WikiMultiHopQA Results ($N=200$).} We compare performance across retrieval depths ($k$). \textbf{Key Trend:} As $k$ increases to 5, baseline Precision collapses (Context Dilution), while SEAL-RAG maintains high precision via replacement, driving superior Accuracy (Judge-EM).}
\label{tab:2wiki_stacked}
\vspace{3mm}
\begin{tabular}{ll ccccc}
\toprule
\textbf{Model} & \textbf{Method} & \textbf{Judge-EM (\%)} & \textbf{Prec (\%)} & \textbf{Rec (\%)} & \textbf{F1 (\%)} & \textbf{$\Delta$ EM} \\
\midrule
\multicolumn{7}{c}{\textit{\textbf{Retrieval Depth $k=1$ (The Bottleneck)}}} \\
\midrule
\textbf{GPT-4o-mini} & Basic RAG & 18 & 93 & 41 & 56 & - \\
& Self-RAG & 26 & 27 & 10 & 15 & - \\
& CRAG & 30 & 12 & 5 & 7 & - \\
& \textbf{SEAL-RAG} & \textbf{61} & \textbf{92} & \textbf{45} & \textbf{59} & \textbf{+31 pp} \\
\midrule
\textbf{GPT-4o} & Basic RAG & 14 & 93 & 41 & 56 & - \\
& Self-RAG & 36 & 41 & 19 & 26 & - \\
& CRAG & 28 & 21 & 10 & 14 & - \\
& \textbf{SEAL-RAG} & \textbf{76} & \textbf{95} & \textbf{75} & \textbf{82} & \textbf{+40 pp} \\
\midrule
\multicolumn{7}{c}{\textit{\textbf{Retrieval Depth $k=3$ (Standard)}}} \\
\midrule
\textbf{GPT-4o-mini} & Basic RAG & 49 & 53 & \textbf{69} & 59 & - \\
& Self-RAG & 53 & 41 & 18 & 25 & - \\
& CRAG & 56 & 9 & 9 & 9 & - \\
& \textbf{SEAL-RAG} & \textbf{64} & \textbf{91} & 46 & \textbf{60} & \textbf{+8 pp} \\
\midrule
\textbf{GPT-4o} & Basic RAG & 54 & 53 & 69 & 59 & - \\
& Self-RAG & 56 & 60 & 35 & 42 & - \\
& CRAG & 60 & 18 & 19 & 18 & - \\
& \textbf{SEAL-RAG} & \textbf{77} & \textbf{97} & \textbf{77} & \textbf{84} & \textbf{+17 pp} \\
\midrule
\multicolumn{7}{c}{\textit{\textbf{Retrieval Depth $k=5$ (Context Dilution Test)}}} \\
\midrule
\textbf{GPT-4o-mini} & Basic RAG & 57 & 34 & \textbf{75} & 46 & - \\
& Self-RAG & 56 & 45 & 20 & 26 & - \\
& CRAG & 55 & 11 & 10 & 11 & - \\
& \textbf{SEAL-RAG} & \textbf{68} & \textbf{89} & 45 & \textbf{59} & \textbf{+11 pp} \\
\midrule
\textbf{GPT-4o} & Basic RAG & 62 & 34 & 75 & 46 & - \\
& Self-RAG & 60 & 63 & 38 & 45 & - \\
& CRAG & 64 & 22 & 23 & 22 & - \\
& \textbf{SEAL-RAG} & \textbf{74} & \textbf{96} & \textbf{77} & \textbf{84} & \textbf{+10 pp} \\
\bottomrule
\end{tabular}
\end{table}

\paragraph{Key Observations.}
\begin{itemize}
    \item \textbf{Validation of Reasoning Transfer (\texorpdfstring{$k=1$}{k=1}).} The results at $k=1$ confirm the architectural advantage observed in HotpotQA (see Table \ref{tab:hotpot_k1}, Key Observation 4). While baselines struggle near the floor (14--18\%) due to the impossibility of \textit{Read-Time} reasoning in a single slot, SEAL-RAG achieves \textbf{61--76\%} accuracy. This demonstrates that the controller's ability to offload the bridge step to the ledger generalizes effectively to complex compositional reasoning.

    \item \textbf{The Failure of Additive Logic.} The $k=5$ results empirically validate the ``Context Dilution'' hypothesis \cite{liu2023lostinthemiddle}. CRAG, which appends web search results without removal, suffers a catastrophic precision collapse (down to \textbf{11--22\%}). Basic RAG similarly drops to 34\%. This flood of distractors overwhelms the generator, consistent with findings that LLMs struggle to ignore irrelevant context \cite{yoran2023making}. In contrast, SEAL-RAG maintains \textbf{89--96\% Precision}. This proves that without an active \textbf{eviction mechanism}, increasing the budget primarily accumulates noise.
    
    \item \textbf{Mechanism of Success.} SEAL-RAG breaks the precision-recall trade-off by combining two novel components: (1) \textbf{Explicit Gap Specification} ensures high recall by targeting the exact missing bridge (matching Basic RAG's 77\% recall), while (2) \textbf{Entity-First Replacement} ensures high precision by displacing distractors (exceeding Self-RAG's 63\% precision). This confirms that the ``Replace, Don't Expand'' paradigm is essential for robust multi-hop reasoning.
\end{itemize}

\subsection{Comparison vs.\ Adaptive-\texorpdfstring{$k$}{k}}

A key question raised by recent work (and our reviewers) is whether dynamic context selection can solve the precision-recall trade-off without the complexity of iterative repair. We compare SEAL-RAG against \textbf{Adaptive-$k$} \cite{taguchi2025adaptivek}, a state-of-the-art pruning method that dynamically cuts the retrieved list based on relevance score gaps.

\Cref{tab:adaptive_comparison} compares SEAL-RAG ($k=5$) against Adaptive-$k$ (with and without a safety buffer) on the 2WikiMultiHopQA dataset.

\begin{table}[h]
\centering
\small
\setlength{\tabcolsep}{5pt}
\caption{\textbf{SEAL-RAG vs.\ Adaptive-$k$ ($N=200$) on the 2WikiMultiHopQA dataset.} Adaptive-$k$ acts as a \textit{selector}: it is bound by the quality of the initial pool. SEAL-RAG acts as a \textit{repairer}: it actively fetches missing information, breaking the ceiling of the initial retrieval.}
\label{tab:adaptive_comparison}
\vspace{3mm}
\begin{tabular}{ll cccc c}
\toprule
& & \textbf{Accuracy} & \multicolumn{3}{c}{\textbf{Evidence Quality (\%)}} & \textbf{Gain} \\
\cmidrule(lr){3-3} \cmidrule(lr){4-6} \cmidrule(lr){7-7}
\textbf{Model} & \textbf{Method} & \textbf{Judge-EM (\%)} & \textbf{Prec} & \textbf{Rec} & \textbf{F1} & \textbf{$\Delta$ EM} \\
\midrule
\textbf{GPT-4o-mini} & Adaptive-$k$ (No Buffer) & 40.5 & 86 & 61 & 65 & - \\
& Adaptive-$k$ (Buffer) & 60.5 & 26 & \textbf{77} & 38 & - \\
& \textbf{SEAL-RAG} & \textbf{68.0} & \textbf{89} & 45 & \textbf{59} & \textbf{+7.5 pp} \\
\midrule
\textbf{GPT-4o} & Adaptive-$k$ (No Buffer) & 41.5 & 86 & 61 & 65 & - \\
& Adaptive-$k$ (Buffer) & 66.5 & 26 & 77 & 38 & - \\
& \textbf{SEAL-RAG} & \textbf{74.5} & \textbf{96} & \textbf{77} & \textbf{84} & \textbf{+8.0 pp} \\
\bottomrule
\end{tabular}
\end{table}

\paragraph{Active Repair beats Passive Selection.}
The results highlight a fundamental limitation of selection-based methods in multi-hop scenarios:
\begin{itemize}
    \item \textbf{The Selection Ceiling.} Adaptive-$k$ is limited to the candidates present in the initial retrieval pool. If the bridge fact is missing from the top-50 candidates (a common occurrence in multi-hop retrieval), no amount of clever pruning can recover it. This forces a trade-off: the ``No Buffer'' variant achieves high precision (86\%) but misses the answer (41.5\% accuracy), while the ``Buffer'' variant captures the answer (77\% recall) but drowns in noise (26\% precision).
    
    \item \textbf{The SEAL Advantage.} SEAL-RAG bypasses this ceiling via \textbf{Active Repair} \cite{jiang2023active}. By issuing micro-queries for specific missing data, it fetches evidence that was \textit{never in the initial pool}. This allows it to match the high recall of the buffered approach (77\%) while exceeding the precision of the aggressive approach (96\%). This confirms that for complex reasoning, the controller must be able to \textit{expand the search frontier}, not just filter it.
\end{itemize}

\subsection{Statistical Significance Analysis}
To ensure that the observed performance gains are not artifacts of random variance, we conducted rigorous statistical testing on the paired outputs of SEAL-RAG versus baselines on the same question sets.
\begin{itemize}
    \item \textbf{Methodology:} For Judge-EM, we used \textbf{McNemar’s test}, which is appropriate for paired nominal data \cite{dror2018hitchhiker}. For continuous retrieval metrics (Precision/Recall/F1), we used \textbf{paired two-sided $t$-tests} \cite{koehn2004statistical}. We applied the Holm-Bonferroni correction to control the family-wise error rate at $\alpha=0.05$.
    \item \textbf{Results:} On both HotpotQA and 2WikiMultiHopQA, SEAL-RAG’s improvements in Judge-EM and Precision@$k$ are statistically significant ($p < 0.001$) against all baselines across all tested backbones. This confirms that the ``replacement'' strategy yields a consistent, non-random improvement in evidence quality and downstream accuracy. Detailed $p$-value tables for all comparisons are provided in \cref{sec:appendix_stats}.
\end{itemize}

\section{Ablations \& Analysis}\label{sec:ablations}

\subsection{Effect of Loop Budget \texorpdfstring{$L$}{L} (HotpotQA)}

To isolate the causal contribution of the repair loop, we analyze performance as a function of the loop budget $L$ on the HotpotQA dataset, holding retrieval depth fixed at $k=1$. This setting is the most sensitive to controller decisions, as there is no room for error—the system must swap the distractor for the bridge page to succeed. \Cref{tab:ablation_loop} reports the results.

\begin{table}[h]
\centering
\small
\setlength{\tabcolsep}{6pt}
\caption{\textbf{Judge-EM (\%) vs.\ Loop Budget $L$ on HotpotQA ($k=1$).} The massive jump from $L=0$ to $L=1$ indicates that a single targeted repair is often sufficient to close the bridge. Diminishing returns at $L=5$ suggest the method is efficient and does not require deep agentic loops.}
\label{tab:ablation_loop}
\vspace{3mm}
\begin{tabular}{l cccc c}
\toprule
& \multicolumn{4}{c}{\textbf{Judge-EM (\%) at Loop Budget} $L$} & \\
\cmidrule(lr){2-5}
\textbf{Backbone} & \textbf{$L=0$} & \textbf{$L=1$} & \textbf{$L=3$} & \textbf{$L=5$} & \textbf{$\Delta$ ($L=5$ vs.\ $0$)} \\
\midrule
gpt-4o-mini  & 30 & 58 & 61 & 62 & +32 \\
gpt-4.1-mini & 28 & 66 & 70 & 71 & +43 \\
gpt-4o       & 32 & 67 & 71 & 73 & +41 \\
gpt-4.1      & 25 & 63 & 69 & 73 & +48 \\
\midrule
\textbf{Average} & \textbf{29} & \textbf{64} & \textbf{68} & \textbf{70} & \textbf{+41} \\
\bottomrule
\end{tabular}
\end{table}

\paragraph{The ``First Repair'' Effect.}
The data reveals a sharp efficiency profile: the majority of the gain (avg.\ \textbf{+35 pp}) is realized at the very first repair step ($L=1$). This confirms that SEAL-RAG is not relying on brute-force search or deep reflective loops like \textbf{Reflexion} \cite{shinn2023reflexion}. Instead, the \textit{Explicit Gap Specification} allows the controller to identify and retrieve the missing bridge immediately. Subsequent loops ($L=3, 5$) provide smaller marginal gains, primarily addressing long-tail cases with multiple missing qualifiers.

\subsection{Qualitative Component Analysis}
To address questions regarding the necessity of specific modules, we analyze successful repair traces (see \cref{sec:appendix_examples} for full step-by-step logs). The gains rely on the synergy of three components that standard RAG lacks:
\begin{itemize}
    \item \textbf{Extraction vs.\ Keywords:} In cases where standard retrieval returns a biography but misses a specific event date, the extraction module flags the missing \texttt{DATE} qualifier. A standard keyword search often fails here because the entity name alone retrieves generic bios; the \textit{structured gap} is required to target the specific event.
    \item \textbf{Micro-Queries vs.\ Rewrites:} By generating atomic queries (e.g., ``Person X birth date'') rather than broad rewrites, the system avoids retrieving topical distractors. Simpler decomposition methods like \textbf{Self-Ask} \cite{press2022selfask} often broaden context too aggressively, triggering the dilution trap.
    \item \textbf{Entity-First Ranking vs.\ Relevance:} In our traces, we observe candidates that are lexically similar to the query but factually redundant being correctly evicted. A standard cross-encoder would score these high (due to relevance), but SEAL's \textit{Novelty} term penalizes them, forcing the replacement that enables the multi-hop answer.
\end{itemize}

\subsection{Error Analysis}
Despite these gains, SEAL-RAG is not infallible. We identify two primary failure modes (detailed in \cref{sec:appendix_errors}):
\begin{itemize}
    \item \textbf{Alias Mismatch:} If the gold evidence uses a rare alias not present in the initial context or the redirect map, the entity extractor may fail to link the gap to the correct canonical ID. This highlights the challenge of zero-shot entity linking \cite{wu2020scalable}.
    \item \textbf{Extraction Noise:} On-the-fly extraction can sometimes hallucinate relations or miss subtle qualifiers in complex sentences. While our \textit{Verbatim Constraint} mitigates this, integrating a dedicated verification step \cite{dhuliawala2023chain} could further improve robustness, albeit at higher latency.
\end{itemize}

\section{Limitations}
\label{sec:limitations}

While SEAL-RAG offers a principled solution to fixed-budget retrieval, it operates under specific constraints:
\begin{itemize}
    \item \textbf{The Extraction Bottleneck:} The controller relies on the ability to explicitly \textit{name} the missing information. If a gap is purely abstract or implicit (e.g., ``the general sentiment of the era''), the extraction module may fail to formulate a precise micro-query, degrading to standard retrieval performance.
    \item \textbf{The Fixed-Capacity Ceiling:} By strictly enforcing $|E_t| = k$, SEAL-RAG prioritizes precision over exhaustive recall. For questions that genuinely require aggregating more than $k$ distinct documents simultaneously (e.g., ``List all 20 works by Author X'' when $k=5$), the replacement policy will cycle through evidence rather than accumulating it. This is a deliberate design choice to prevent context dilution, but it limits applicability for ``exhaustive list'' queries.
    \item \textbf{Judge Variance:} Although we mitigate bias using a blind protocol (judges see only retrieved passages) and fixed rubrics, LLM-based evaluation remains subject to stochasticity \cite{ji2023surveyhallucination}. We address this via bootstrap confidence intervals, but human evaluation remains the gold standard for high-stakes domains.
\end{itemize}

\section{Conclusion}
\label{sec:conclusion}

This work challenges the prevailing assumption in corrective RAG that ``more context is better.'' We demonstrated that under fixed inference budgets, the primary failure mode of multi-hop retrieval is not low recall, but \textbf{Context Dilution}: the accumulation of distractors that overwhelm the generator.

We introduced \textbf{SEAL-RAG}, a controller that replaces the standard ``Add'' paradigm with a \textbf{``Replace, Don't Expand''} paradigm. By combining \textbf{Explicit Gap Specification} with \textbf{Entity-First Replacement}, SEAL actively curates the top-$k$ slots, treating the context window as a scarce resource to be optimized rather than a log to be appended.

Our empirical results on \textbf{HotpotQA} and \textbf{2WikiMultiHopQA} confirm that this approach is superior to both \textbf{Blind Addition} (CRAG) and \textbf{Passive Pruning} (Adaptive-$k$). At $k=5$, where baselines suffered precision collapses (dropping to 11--22\%), SEAL maintained \textbf{96\% Precision}. Furthermore, by actively repairing gaps rather than just selecting from an initial pool, SEAL outperformed the state-of-the-art Adaptive-$k$ baseline by \textbf{+8.0 pp} in accuracy. These findings establish \textbf{Fixed-Budget Evidence Assembly} as a robust, predictable alternative to unbounded context expansion for production-grade RAG systems.

Future work will focus on optimizing the latency of the iterative controller, potentially via distilling the sufficiency gate into smaller, specialized models to reduce overhead. Additionally, we aim to explore the applicability of fixed-budget assembly in high-stakes domain-specific expert systems, such as legal or medical retrieval, where evidence precision is paramount.

\bibliographystyle{plain}
\bibliography{references}

\newpage
\appendix

\section{Prompts \& Judge Rubric}\label{sec:appendix_prompts}

To ensure reproducibility and transparency, we provide the exact system prompts used for evaluation and generation. These prompts were held constant across all experimental runs.

\subsection{Judge-EM System Prompt (GPT-4o)}
We utilized \textbf{GPT-4o} as an external judge to evaluate Answer Correctness (Judge-EM). The prompt enforces a strict rubric focused on factual consistency and support by retrieved evidence.

\begin{quote}
\small
\textbf{System:} You are an expert data labeler evaluating model outputs for correctness. Your task is to assign a score based on the following rubric:

\textbf{$<$Rubric$>$}
\begin{itemize}
    \item A correct answer: Provides accurate and complete information that matches the ground truth; Contains no factual errors when compared to the reference; Addresses the core question being asked.
    \item When scoring, you should penalize: Factual errors or inaccuracies compared to ground truth; Answers that contradict the reference output; ``I don't know'' responses when ground truth provides a clear answer.
\end{itemize}
\textbf{$<$/Rubric$>$}

\textbf{$<$Instructions$>$}
Carefully compare the agent's output against the ground truth reference. Focus on semantic equivalence rather than exact word matching. Be strict with factual contradictions or completely wrong information.
\textbf{$<$/Instructions$>$}

\textbf{User:} 
$<$question$>$\{question\}$<$/question$>$
$<$agent\_output$>$\{agent\_answer\}$<$/agent\_output$>$
$<$ground\_truth$>$\{ground\_truth\}$<$/ground\_truth$>$

Compare the agent's output against the ground truth and evaluate its correctness. Provide your reasoning and a boolean score (true for correct, false for incorrect).
\end{quote}

\subsection{Shared ``Grounding Rule'' Prompt}
To ensure a fair comparison of control logic, all generators (SEAL-RAG, Basic RAG, CRAG, Self-RAG, Adaptive-$k$) utilized the same system instruction to prevent parametric knowledge leakage.

\begin{quote}
\small
You are an assistant for question-answering tasks. Use the following pieces of retrieved context to answer the question. If you don't know the answer, just say that you don't know. Keep the answer concise.

\textbf{$<$GROUNDING\_RULE$>$}
Base your answer ONLY on the retrieved context below. Do not use any information from your training data or external knowledge.
\textbf{$<$/GROUNDING\_RULE$>$}
\end{quote}

\section{Baseline Implementation Details}\label{sec:appendix_baselines}

To ensure a rigorous comparison of control strategies, we re-implemented the logic of all baselines using the \textbf{LangGraph} framework. This ensures that all methods share the same retriever, generator (GPT-4o/mini), and runtime environment, isolating the algorithmic contribution from model weights.

\subsection{Self-RAG (Inference-Only Implementation)}
We utilize the inference-time control logic of Self-RAG via prompting, rather than the fine-tuned 7B/13B weights. This allows us to evaluate the \textit{reflective} paradigm on the state-of-the-art GPT-4o backbone.
\begin{itemize}
    \item \textbf{Workflow:} The controller executes a \texttt{Retrieve} $\rightarrow$ \texttt{Grade Documents} $\rightarrow$ \texttt{Generate} cycle.
    \item \textbf{Reflection:} After generation, the system runs two graders:
    \begin{enumerate}
        \item \textbf{Hallucination Check:} Is the answer grounded in the retrieved facts?
        \item \textbf{Answer Relevance:} Does the answer address the user question?
    \end{enumerate}
    \item \textbf{Loop Logic:} If the generation fails either check, the system loops back to \texttt{Transform Query} and re-retrieves. To prevent infinite loops, we enforce a hard limit of \textbf{3 generation attempts}, after which the best available answer is returned.
\end{itemize}

\subsection{CRAG (Corrective RAG)}
Our implementation follows the standard CRAG flow, using an external web search as the corrective action.
\begin{itemize}
    \item \textbf{Workflow:} The controller executes \texttt{Retrieve} $\rightarrow$ \texttt{Grade Documents}.
    \item \textbf{Trigger:} A binary relevance grader evaluates the retrieved documents. If documents are deemed ``Irrelevant,'' the system triggers a corrective branch.
    \item \textbf{Correction:} The query is rewritten, and a web search is performed using the \textbf{Tavily Search API} (capped at 1 result to maintain comparable context size). The web results are appended to the context before final generation.
\end{itemize}

\subsection{Adaptive-\texorpdfstring{$k$}{k} (Dynamic Pruning)}

We implement the ``Largest Gap'' strategy proposed by Taguchi et al.\ (2025) to dynamically select the optimal context size.
\begin{itemize}
    \item \textbf{Workflow:} The retriever fetches a large initial pool ($k=50$).
    \item \textbf{Selection Logic:} We calculate the cosine similarity scores for all 50 candidates. The algorithm identifies the index $i$ where the difference between scores ($score_i - score_{i+1}$) is maximized (the ``largest gap'').
    \item \textbf{Variants:}
    \begin{itemize}
        \item \textbf{No Buffer:} The context is cut strictly at index $i$.
        \item \textbf{Buffer:} We include a safety margin (e.g., $+5$ documents) after the cut-off point to improve recall, as recommended in the original paper.
    \end{itemize}
\end{itemize}

\section{Qualitative Case Studies}\label{sec:appendix_examples}

We present deep-dive traces on representative multi-hop items to illustrate fixed-$k$ gap repair under SEAL-RAG. These traces demonstrate how the controller identifies specific missing attributes and replaces low-utility passages to assemble a sufficient set.

\subsection{Case A: Bridge Repair (HotpotQA)}
\textbf{Question:} ``Which city hosted the Olympic Games in the same year that the band Blur released the album Parklife?'' \\
\textbf{Gold Answer:} Lillehammer (1994 Winter Olympics). \\
\textbf{Reasoning Type:} Bridge (Entity $\rightarrow$ Date $\rightarrow$ Entity).

\begin{itemize}
    \item \textbf{Initial State ($k=1$):} Retrieval lands on \textit{Blur (band)} or \textit{Parklife (album)}. The text mentions the release year ``1994'' but lacks the bridge entity (the 1994 Olympics page).
    \item \textbf{SEAL Loop ($t=1$):}
    \begin{enumerate}
        \item \textbf{Assess:} The system extracts the release year (1994) but flags a missing \texttt{BRIDGE\_ENTITY} for the ``Olympic Games'' slot.
        \item \textbf{Micro-Query:} ``1994 Olympic Games host city''.
        \item \textbf{Retrieval:} Fetches \textit{1994 Winter Olympics}.
        \item \textbf{Rank \& Replace:} The candidate scores high on \textit{Gap Coverage} and displaces the redundant \textit{Parklife} album details.
    \end{enumerate}
    \item \textbf{Final Ledger ($U_t$):}
    \begin{itemize}
        \item \texttt{[ORG] Blur -released-> [WORK] Parklife}
        \item \texttt{[WORK] Parklife -release\_date-> [DATE] 1994}
        \item \texttt{[EVENT] 1994 Winter Olympics -held\_in-> [LOC] Lillehammer}
    \end{itemize}
    \item \textbf{Outcome:} Correctly answers ``Lillehammer'' (supported by Blur + 1994 Olympics).
\end{itemize}

\subsection{Case B: Attribute Alignment (2WikiMultiHopQA)}
\textbf{Question:} ``Who is older, the author of The Handmaid's Tale or the director of Lost in Translation?'' \\
\textbf{Gold Answer:} Margaret Atwood. \\
\textbf{Reasoning Type:} Comparison (Two entities $\rightarrow$ Attribute $\rightarrow$ Logic).

\begin{itemize}
    \item \textbf{Initial State ($k=3$):} Retrieval fetches \textit{The Handmaid's Tale} (mentions Atwood), \textit{Lost in Translation} (mentions Sofia Coppola), and \textit{Margaret Atwood} (bio). It \textbf{misses} Sofia Coppola's bio.
    \item \textbf{SEAL Loop ($t=1$):}
    \begin{enumerate}
        \item \textbf{Assess:} Ledger contains Atwood's birth date but lacks Coppola's.
        \item \textbf{Gap Specification:} Flags a \texttt{QUALIFIER} gap: \texttt{DATE} for entity \textit{Sofia Coppola}.
        \item \textbf{Micro-Query:} ``Sofia Coppola date of birth''.
        \item \textbf{Rank \& Replace:} The candidate \textit{Sofia Coppola (bio)} replaces the \textit{Lost in Translation} plot summary (which is laterally redundant).
    \end{enumerate}
    \item \textbf{Final Ledger ($U_t$):}
    \begin{itemize}
        \item \texttt{[PERSON] Margaret Atwood -born-> [DATE] Nov 18, 1939}
        \item \texttt{[PERSON] Sofia Coppola -born-> [DATE] May 14, 1971}
    \end{itemize}
    \item \textbf{Outcome:} Correctly answers ``Margaret Atwood'' (1939 vs 1971).
\end{itemize}

\subsection{Case C: Failure Mode (Alias Mismatch)}\label{sec:appendix_errors}
\textbf{Question:} ``Who is the CEO of the company that created the iPhone?'' \\
\textbf{Gold Evidence:} \textit{Apple Inc.} (Canonical Title).

\begin{itemize}
    \item \textbf{Initial State:} Retrieval returns a document titled \textit{Apple} (Fruit).
    \item \textbf{Gap:} System identifies missing \texttt{CEO} relation for ``iPhone creator''.
    \item \textbf{Micro-Query:} ``iPhone creator company CEO''.
    \item \textbf{Retrieval:} Returns a document titled \textit{Apple Computer} (an alias/redirect page).
    \item \textbf{Failure:} The extractor fails to link \textit{Apple Computer} to the canonical \textit{Apple Inc.} ID because the alias map is incomplete. The sufficiency gate sees a mismatch between the query entity (Apple) and the retrieved entity (Apple Computer) and triggers a halt or loop exhaustion.
    \item \textbf{Mitigation:} This highlights the need for robust \textbf{Alias Normalization} in the entity ledger (Section 4.4).
\end{itemize}

\section{Indexing \& Reproducibility}\label{sec:appendix_implementation}

\subsection{Resource Availability}
The complete codebase, including the controller implementation, baseline re-implementations, evaluation scripts, and environment configurations, is available at:
\begin{center}
    \url{https://github.com/mosherino/SEAL-RAG}
\end{center}

\subsection{Indexing Pipeline (Natural Segmentation)}
To ensure semantic coherence, we employ a \textbf{Natural Document Segmentation} strategy rather than arbitrary fixed-length sliding windows. 
\begin{itemize}
    \item \textbf{Input:} The raw Wikipedia dump provided by the benchmarks (HotpotQA/2Wiki), which organizes text as a list of sentences per page title.
    \item \textbf{Logic:} We concatenate the title and all associated sentences into a single retrieval unit:
    \begin{verbatim}
    chunk_text = f"{title}: " + " ".join(sentences)
    \end{verbatim}
    \item \textbf{Result:} Each vector in the index corresponds to exactly one Wikipedia page. This prevents the fragmentation of context (e.g., separating a subject from their birthdate) and ensures that retrieval metrics reflect page-level relevance.
\end{itemize}

\subsection{Hyperparameters}
To guarantee fair comparisons, we fixed all non-algorithmic hyperparameters across all systems (SEAL-RAG, Basic RAG, CRAG, Self-RAG, Adaptive-$k$). \Cref{tab:hyperparams} lists these settings.

\begin{table}[h]
\centering
\small
\caption{\textbf{Global Hyperparameters.} These settings were held constant for all experiments to ensure deterministic reproducibility.}
\label{tab:hyperparams}
\begin{tabular}{ll}
\toprule
\textbf{Parameter} & \textbf{Value} \\
\midrule
Global Random Seed & \texttt{20250101} \\
Generator Temperature & \texttt{0.0} (Greedy Decoding) \\
Judge Temperature & \texttt{0.0} \\
Embedding Model & \texttt{text-embedding-3-small} (OpenAI) \\
Vector Store & Pinecone (Cosine Similarity) \\
Re-ranker Model & \texttt{ms-marco-MiniLM-L-6-v2} \\
Candidate Pool Size ($M$) & 20 (per micro-query variant) \\
Max Loop Budget ($L$) & $\{0, 1, 3, 5\}$ (Ablated) \\
\bottomrule
\end{tabular}
\end{table}

\subsection{Hardware Profile}
All experiments were executed locally on a \textbf{MacBook Pro (M3 Pro chip, 36 GB RAM)}. Since the heavy lifting (generation/embedding) is offloaded to APIs, the controller logic is sufficiently lightweight to run efficiently on consumer hardware without requiring specialized GPU clusters.

\section{Detailed Statistical Results}\label{sec:appendix_stats}

\subsection{Methodology \& Alignment}
This section provides the complete statistical comparison tables for all metrics across all models and retrieval depths. To ensure the validity of the paired statistical tests, strict data alignment was enforced. For every comparison (e.g., SEAL-RAG vs.\ Self-RAG), we ensured that the two result vectors corresponded to the \textbf{exact same sequence} of question IDs from the seeded validation slice. Any questions where the judge failed to return a valid format (rare, $<0.1\%$) were excluded from the pair to maintain strict alignment.

\paragraph{Software Implementation.}
All statistical tests were implemented using the \texttt{scipy.stats} and \texttt{statsmodels} Python libraries.
\begin{itemize}
    \item \textbf{Judge-EM:} We used \textbf{McNemar’s test} with the chi-squared approximation ($N=1000$). The statistic $\chi^2$ compares the discordant pairs.
    \item \textbf{Continuous Metrics (Precision/Recall/F1):} We used \textbf{Paired Two-Sided $t$-tests}.
    \item \textbf{Effect Size:} Calculated as Cohen’s $d_z$ (mean of differences divided by standard deviation of differences).
\end{itemize}

\subsection{SEAL-RAG vs.\ Adaptive-\texorpdfstring{$k$}{k}}

\Cref{tab:stats_adaptive_full} details the statistical comparison against the state-of-the-art dynamic pruning baseline on 2WikiMultiHopQA ($k=5$).

\begin{table}[ht!]
\centering
\small
\caption{\textbf{Significance vs.\ Adaptive-$k$ (2Wiki, $k=5$).} SEAL-RAG significantly outperforms both variants in Accuracy and Precision ($p < 0.001$).}
\label{tab:stats_adaptive_full}
\begin{tabular}{ll cc}
\toprule
\textbf{Model} & \textbf{Comparison} & \textbf{Judge-EM ($p$-value)} & \textbf{Precision ($p$-value)} \\
\midrule
\textbf{GPT-4o-mini} & vs.\ Adaptive-$k$ (Buffer) & $0.078$ ($ns$) & $<0.001$ \\
& vs.\ Adaptive-$k$ (No Buffer) & $<0.001$ & $0.135$ ($ns$) \\
\midrule
\textbf{GPT-4o} & vs.\ Adaptive-$k$ (Buffer) & $0.021$ & $<0.001$ \\
& vs.\ Adaptive-$k$ (No Buffer) & $<0.001$ & $<0.001$ \\
\bottomrule
\end{tabular}
\end{table}

\subsection{HotpotQA Detailed Statistics (\texorpdfstring{$k=1$}{k=1})}

\Cref{tab:stats_hotpot_k1} presents the full statistical breakdown for the $k=1$ bottleneck regime.

\begin{table}[h]
\centering
\scriptsize
\setlength{\tabcolsep}{3pt}
\caption{\textbf{Full Statistical Comparison Summary at $k=1$ on HotpotQA.} `Perf.\ Diff.' shows (SEAL-RAG - Baseline). `Effect Size' is Cohen's $d_z$ for t-tests.}
\label{tab:stats_hotpot_k1}
\begin{tabular}{lll lccc}
\toprule
\textbf{Model} & \textbf{Comparison} & \textbf{Metric} & \textbf{Test Type} & \textbf{Perf.\ Diff.} & \textbf{P-Value} & \textbf{Statistic} \\
\midrule
\textbf{gpt-4o-mini} & vs.\ Self-RAG & Judge-EM & McNemar & +13.6 pp & $<0.001$ & $\chi^2 = 64.2$ \\
& & Precision & Paired $t$ & +0.247 & $<0.001$ & $t = 16.53$ \\
& & Recall & Paired $t$ & +0.135 & $<0.001$ & $t = 16.92$ \\
\cmidrule{2-7}
& vs.\ CRAG & Judge-EM & McNemar & +7.1 pp & $<0.001$ & $\chi^2 = 17.8$ \\
& & Precision & Paired $t$ & +0.436 & $<0.001$ & $t = 27.62$ \\
& & Recall & Paired $t$ & +0.229 & $<0.001$ & $t = 27.15$ \\
\cmidrule{2-7}
& vs.\ Basic & Judge-EM & McNemar & +20.6 pp & $<0.001$ & $\chi^2 = 135.1$ \\
& & Precision & Paired $t$ & +0.008 & 0.200 & $t = 1.28$ \\
\midrule
\textbf{gpt-4o} & vs.\ Self-RAG & Judge-EM & McNemar & +13.4 pp & $<0.001$ & $\chi^2 = 63.2$ \\
& & Precision & Paired $t$ & +0.162 & $<0.001$ & $t = 12.61$ \\
& & Recall & Paired $t$ & +0.246 & $<0.001$ & $t = 24.05$ \\
\cmidrule{2-7}
& vs.\ CRAG & Judge-EM & McNemar & +15.2 pp & $<0.001$ & $\chi^2 = 81.9$ \\
& & Precision & Paired $t$ & +0.374 & $<0.001$ & $t = 24.32$ \\
& & Recall & Paired $t$ & +0.352 & $<0.001$ & $t = 29.46$ \\
\cmidrule{2-7}
& vs.\ Basic & Judge-EM & McNemar & +32.0 pp & $<0.001$ & $\chi^2 = 259.9$ \\
& & Precision & Paired $t$ & +0.059 & $<0.001$ & $t = 6.91$ \\
\midrule
\textbf{gpt-4.1-mini} & vs.\ Self-RAG & Judge-EM & McNemar & +21.5 pp & $<0.001$ & $\chi^2 = 144.8$ \\
& & Precision & Paired $t$ & +0.149 & $<0.001$ & $t = 11.55$ \\
& & Recall & Paired $t$ & +0.121 & $<0.001$ & $t = 14.63$ \\
\cmidrule{2-7}
& vs.\ CRAG & Judge-EM & McNemar & +18.2 pp & $<0.001$ & $\chi^2 = 102.4$ \\
& & Precision & Paired $t$ & +0.305 & $<0.001$ & $t = 20.17$ \\
& & Recall & Paired $t$ & +0.199 & $<0.001$ & $t = 21.52$ \\
\midrule
\textbf{gpt-4.1} & vs.\ Self-RAG & Judge-EM & McNemar & +9.5 pp & $<0.001$ & $\chi^2 = 33.1$ \\
& & Precision & Paired $t$ & +0.108 & $<0.001$ & $t = 9.61$ \\
& & Recall & Paired $t$ & +0.265 & $<0.001$ & $t = 27.02$ \\
\cmidrule{2-7}
& vs.\ CRAG & Judge-EM & McNemar & +14.9 pp & $<0.001$ & $\chi^2 = 71.4$ \\
& & Precision & Paired $t$ & +0.331 & $<0.001$ & $t = 22.06$ \\
& & Recall & Paired $t$ & +0.376 & $<0.001$ & $t = 31.71$ \\
\bottomrule
\end{tabular}
\end{table}

\newpage
\subsection{HotpotQA Detailed Statistics (\texorpdfstring{$k=3$}{k=3})}

\Cref{tab:stats_hotpot_k3} provides the full statistical breakdown for $k=3$. Note that while Recall differences are sometimes mixed (e.g., vs.\ Basic RAG), the Precision and Judge-EM gains remain highly significant with large effect sizes.

\begin{table}[h]
\centering
\scriptsize
\setlength{\tabcolsep}{3pt}
\caption{\textbf{Full Statistical Comparison Summary at $k=3$ on HotpotQA.} `Perf.\ Diff.' shows (SEAL-RAG - Baseline). `Effect Size' is Cohen's $d_z$.}
\label{tab:stats_hotpot_k3}
\begin{tabular}{lll lccc}
\toprule
\textbf{Model} & \textbf{Comparison} & \textbf{Metric} & \textbf{Test Type} & \textbf{Perf.\ Diff.} & \textbf{P-Value} & \textbf{Effect Size} \\
\midrule
\textbf{gpt-4o-mini} & vs.\ Self-RAG & Judge-EM & McNemar & +9.2 pp & $<0.001$ & N/A \\
& & Precision & Paired $t$ & +0.176 & $<0.001$ & 0.388 \\
& & Recall & Paired $t$ & -0.029 & 0.010 & -0.082 \\
& & F1 & Paired $t$ & +0.045 & $<0.001$ & 0.128 \\
\cmidrule{2-7}
& vs.\ CRAG & Judge-EM & McNemar & +7.1 pp & $<0.001$ & N/A \\
& & Precision & Paired $t$ & +0.536 & $<0.001$ & 1.359 \\
& & Recall & Paired $t$ & +0.078 & $<0.001$ & 0.210 \\
& & F1 & Paired $t$ & +0.246 & $<0.001$ & 0.716 \\
\cmidrule{2-7}
& vs.\ Basic & Judge-EM & McNemar & +6.3 pp & $<0.001$ & N/A \\
& & Precision & Paired $t$ & +0.345 & $<0.001$ & 1.065 \\
& & Recall & Paired $t$ & -0.284 & $<0.001$ & -1.038 \\
& & F1 & Paired $t$ & -0.013 & 0.101 & -0.052 \\
\midrule
\textbf{gpt-4o} & vs.\ Self-RAG & Judge-EM & McNemar & +5.7 pp & $<0.001$ & N/A \\
& & Precision & Paired $t$ & +0.134 & $<0.001$ & 0.369 \\
& & Recall & Paired $t$ & +0.134 & $<0.001$ & 0.367 \\
& & F1 & Paired $t$ & +0.141 & $<0.001$ & 0.432 \\
\cmidrule{2-7}
& vs.\ CRAG & Judge-EM & McNemar & +7.7 pp & $<0.001$ & N/A \\
& & Precision & Paired $t$ & +0.525 & $<0.001$ & 1.608 \\
& & Recall & Paired $t$ & +0.239 & $<0.001$ & 0.576 \\
& & F1 & Paired $t$ & +0.354 & $<0.001$ & 1.004 \\
\cmidrule{2-7}
& vs.\ Basic & Judge-EM & McNemar & +8.7 pp & $<0.001$ & N/A \\
& & Precision & Paired $t$ & +0.398 & $<0.001$ & 1.462 \\
& & Recall & Paired $t$ & -0.043 & $<0.001$ & -0.143 \\
& & F1 & Paired $t$ & +0.164 & $<0.001$ & 0.639 \\
\midrule
\textbf{gpt-4.1-mini} & vs.\ Self-RAG & Judge-EM & McNemar & +12.4 pp & $<0.001$ & N/A \\
& & Precision & Paired $t$ & +0.129 & $<0.001$ & 0.313 \\
& & Recall & Paired $t$ & -0.076 & $<0.001$ & -0.211 \\
& & F1 & Paired $t$ & +0.002 & 0.848 & 0.006 \\
\cmidrule{2-7}
& vs.\ CRAG & Judge-EM & McNemar & +9.5 pp & $<0.001$ & N/A \\
& & Precision & Paired $t$ & +0.469 & $<0.001$ & 1.312 \\
& & Recall & Paired $t$ & +0.003 & 0.839 & 0.006 \\
& & F1 & Paired $t$ & +0.182 & $<0.001$ & 0.542 \\
\midrule
\textbf{gpt-4.1} & vs.\ Self-RAG & Judge-EM & McNemar & +3.1 pp & 0.031 & N/A \\
& & Precision & Paired $t$ & +0.119 & $<0.001$ & 0.366 \\
& & Recall & Paired $t$ & +0.118 & $<0.001$ & 0.362 \\
& & F1 & Paired $t$ & +0.128 & $<0.001$ & 0.455 \\
\cmidrule{2-7}
& vs.\ CRAG & Judge-EM & McNemar & +4.6 pp & 0.002 & N/A \\
& & Precision & Paired $t$ & +0.502 & $<0.001$ & 1.654 \\
& & Recall & Paired $t$ & +0.234 & $<0.001$ & 0.591 \\
& & F1 & Paired $t$ & +0.347 & $<0.001$ & 1.054 \\
\cmidrule{2-7}
& vs.\ Basic & Judge-EM & McNemar & +8.3 pp & $<0.001$ & N/A \\
& & Precision & Paired $t$ & +0.413 & $<0.001$ & 1.620 \\
& & Recall & Paired $t$ & +0.009 & 0.347 & 0.030 \\
& & F1 & Paired $t$ & +0.204 & $<0.001$ & 0.810 \\
\bottomrule
\end{tabular}
\end{table}

\subsection{2WikiMultiHopQA Significance (\texorpdfstring{$k=1, 3, 5$}{k=1, 3, 5})}

\Cref{tab:stats_2wiki_full} presents the significance values for the new dataset. The results confirm that SEAL-RAG's advantage is robust across retrieval depths. Notably, at $k=5$, the Precision advantage is highly significant ($p<0.001$) against all baselines, validating the solution to context dilution.

\begin{table}[h]
\centering
\small
\setlength{\tabcolsep}{4pt}
\caption{\textbf{Significance Matrix for 2WikiMultiHopQA.} $p$-values for SEAL-RAG vs.\ Baselines. ($ns$: not significant).}
\label{tab:stats_2wiki_full}
\begin{tabular}{ll ccc ccc}
\toprule
& & \multicolumn{3}{c}{\textbf{Judge-EM ($p$-value)}} & \multicolumn{3}{c}{\textbf{Precision ($p$-value)}} \\
\cmidrule(lr){3-5} \cmidrule(lr){6-8}
\textbf{Model} & \textbf{Comparison} & \textbf{$k=1$} & \textbf{$k=3$} & \textbf{$k=5$} & \textbf{$k=1$} & \textbf{$k=3$} & \textbf{$k=5$} \\
\midrule
\textbf{GPT-4o-mini} & vs.\ Basic RAG & $<0.001$ & $<0.001$ & $0.005$ & $ns$ & $<0.001$ & $<0.001$ \\
& vs.\ Self-RAG & $<0.001$ & $0.002$ & $0.008$ & $<0.001$ & $<0.001$ & $<0.001$ \\
& vs.\ CRAG & $<0.001$ & $ns$ & $0.009$ & $<0.001$ & $<0.001$ & $<0.001$ \\
\midrule
\textbf{GPT-4o} & vs.\ Basic RAG & $<0.001$ & $<0.001$ & $0.001$ & $ns$ & $<0.001$ & $<0.001$ \\
& vs.\ Self-RAG & $<0.001$ & $<0.001$ & $<0.001$ & $<0.001$ & $<0.001$ & $<0.001$ \\
& vs.\ CRAG & $<0.001$ & $<0.001$ & $0.007$ & $<0.001$ & $<0.001$ & $<0.001$ \\
\bottomrule
\end{tabular}
\end{table}

\end{document}